\documentclass[11pt,a4paper]{article}
\usepackage{authblk}
\usepackage[hyperref]{acl2019}
\usepackage{times}
\usepackage{latexsym}
\usepackage{arabtex}
\usepackage{amsmath,amssymb,amsfonts}
\usepackage{algorithmic}
\usepackage{graphicx}
\usepackage{textcomp}
\usepackage{xcolor}
\usepackage{utf8}
\usepackage{hyperref}
\usepackage{multirow}

\setcode{utf8}
\usepackage[utf8]{inputenc} 

\renewenvironment{abstract}%
         {\centerline{\large\bf Abstract}%
          \begin{list}{}%
             {\setlength{\rightmargin}{0.6cm}%
              \setlength{\leftmargin}{0.6cm}}%
           \item[]\ignorespaces}%
         {\unskip\end{list}}

\aclfinalcopy % Uncomment this line for the final submission
\title{Tha3aroon at NSURL-2019 Task 8: Semantic Question Similarity in Arabic}

\author{Ali Fadel}
\author{Ibraheem Tuffaha}
\author{Mahmoud Al-Ayyoub}

\affil{Jordan University of Science and Technology, Irbid, Jordan \authorcr
  \{\tt aliosm1997, bro.t.1996, malayyoub\}@gmail.com}

\date{\today}

\begin{document}
\maketitle
\begin{abstract}
In this paper, we describe our team's effort on the semantic text question similarity task of NSURL 2019. Our top performing system utilizes several innovative data augmentation techniques to enlarge the training data. Then, it takes ELMo pre-trained contextual embeddings of the data and feeds them into an ON-LSTM network with self-attention. This results in sequence representation vectors that are used to predict the relation between the question pairs. The model is ranked in the 1st place with 96.499 F1-score (same as the second place F1-score) and the 2nd place with 94.848 F1-score (differs by 1.076 F1-score from the first place) on the public and private leaderboards, respectively.
\end{abstract}

\section{Introduction}

Semantic Text Similarity (STS) problems are both real-life and challenging. For example, in the paraphrase identification task, STS is used to predict if one sentence is a paraphrase of the other or not \cite{madnani2012re,he2015multi,zain2017paraphrase}. Also, in answer sentence selection task, it is utilized to determine the relevance between question-answer pairs and rank the answers sentences from the most relevant to the least. This idea can also be applied to search engines in order to find documents relevant to a query \cite{yang2015wikiqa,tan2018multiway,yang2019xlnet}.

A new task has been proposed by Mawdoo3\footnote{\url{ https://www.mawdoo3.com}} company with a new dataset provided by their data annotation team for Semantic Question Similarity (SQS) for the Arabic language \cite{schwab2017semantic,mahmoud2017text,alian2018arabic}. SQS is a variant of STS, which aims to compare a pair of questions and determine whether they have the same meaning or not.
The SQS in Arabic task is one of the shared tasks of the Workshop on NLP Solutions for Under Resourced Languages (NSURL 2019) and it consists of 12K questions pairs \cite{seelawi-EtAl:2019:NSURL}.

In this paper, we describe our team's efforts to tackle this task. After preprocessing the data, we use four data augmentation steps to enlarge the training data to about four times the size of the original training data. We then build a neural network model with four components. The model uses ELMo (which stands for Embeddings from Language Models) \cite{peters2018deep} pre-trained contextual embeddings as an input and builds sequence representation vectors that are used to predict the relation between the question pairs. The task is hosted on Kaggle\footnote{\url{https://www.kaggle.com}} platform and our model is ranked in the first place with 96.499 F1-score (same as the second place F1-score) and in the second place with 94.848 F1-score (differs by 1.076 F1-score from the first place) on the public and private leaderboards, respectively.

The rest of this paper is organized as follows.
%In Section~\ref{sec:related}, we provide a general literature review for the existing approaches.
In Section~\ref{sec:method}, we describe our methodology, including data preprocessing, data augmentation, and model structure, while in Section~\ref{sec:result}, we present our experimental results and discuss some insights from our model. Finally, the paper is concluded in Section~\ref{sec:conc}.

\section{Methodology}
\label{sec:method}

In this section, we present a detailed description of our model. We start by discussing the preprecessing steps we take before going into the details of the first novel aspect of our work, which is the data augmentation techniques. We then discuss the neural network model starting from the input all the way to the decision step. The implementation is available on a public repository.\footnote{\url{https://github.com/AliOsm/semantic-question-similarity}}

\subsection{Data Preprocessing}

\begin{figure}
    \centering
    \includegraphics[width=0.25\textwidth]{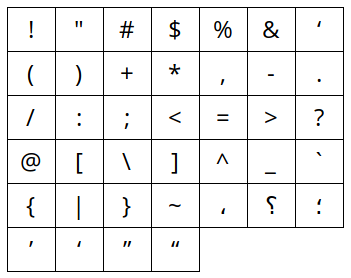}
    \caption{Punctuation marks considered in the preprocessing step}
    \label{punctuations}
\end{figure}

In this work, we only consider one preprocessing step, which is to separate the punctuation marks shown in Figure~\ref{punctuations} from the letters.
For example, if the question was: 
``\<مرحبا، كيف الحال؟>'', then it will be processed as follows:
``\<مرحبا ، كيف الحال ؟>''.
This is done to preserve as much information as possible in the questions while keeping the words clear of punctuations.

\subsection{Data Augmentation}
\label{sec:data_aug}

The training data contains 11,997 question pairs: 5,397 labeled as 1 (i.e., similar) and 6,600 labeled as 0 (i.e., not similar). To obtain a larger dataset, we augment the data using the following rules.

Suppose we have questions A, B and C
\begin{itemize}
  \item \textbf{Positive Transitive}:
  
  If A is similar to B, and B is similar to C, then A is similar to C.
  \item \textbf{Negative Transitive}:
  
  If A is similar to B, and B is \textit{NOT} similar to C, then A is \textit{NOT} similar to C.
  
  \textbf{Note:} The previous two rules generates 5,490 extra examples (bringing the total up to 17,487).
  \item \textbf{Symmetric}:
  
  If A is similar to B then B is similar to A, and if A is not similar to B then B is not similar to A.
  
  \textbf{Note:} This rule doubles the number of examples to 34,974 in total.
  \item \textbf{Reflexive}:
  
  By definition, a question A is similar to itself.
  
  \textbf{Note:} This rule generates 10,540 extra positive examples (45,514 total) which helps balancing the number of positive and negative examples.
\end{itemize}

\begin{figure}
    \centering
    \includegraphics[width=0.48\textwidth]{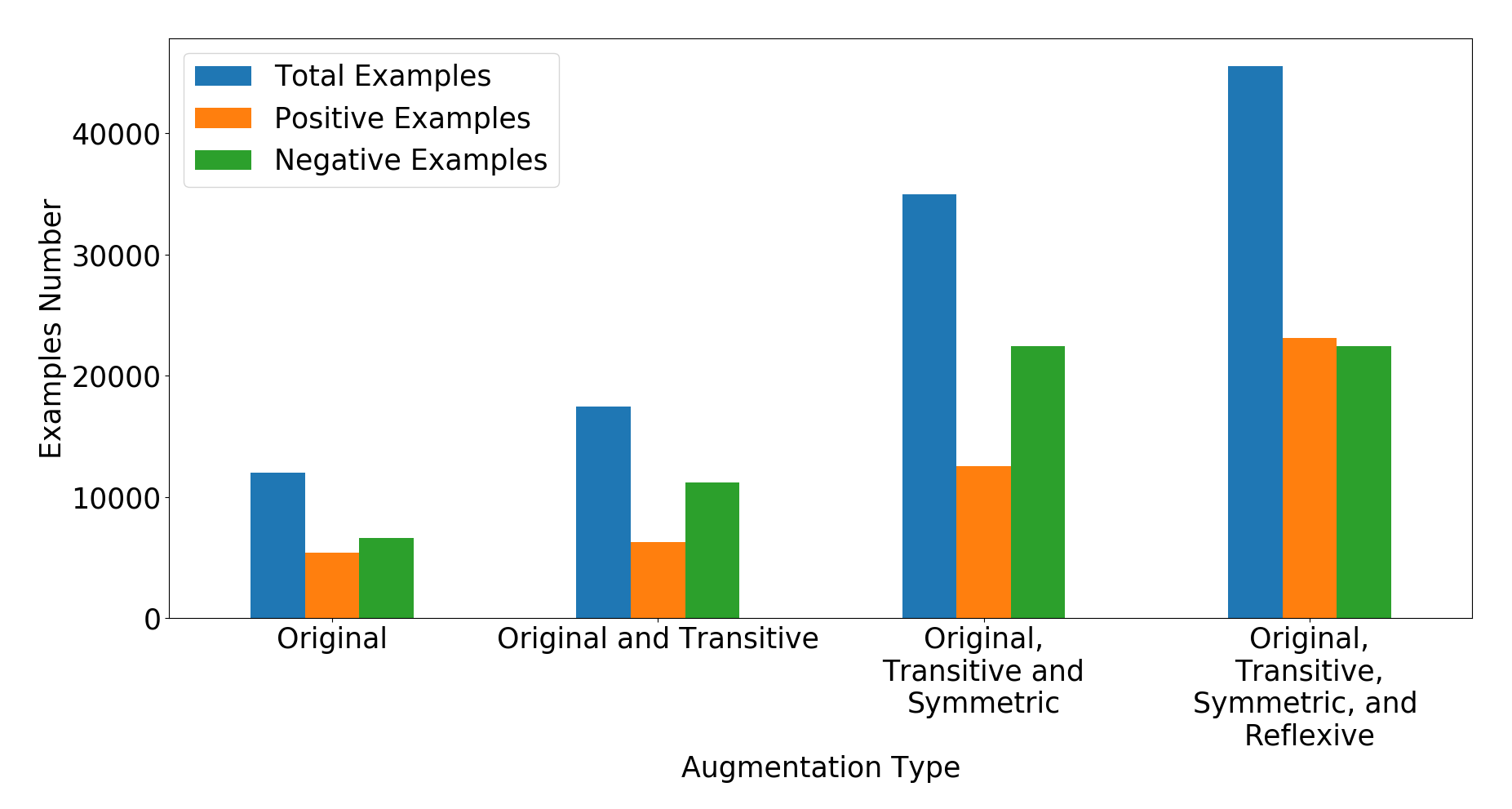}
    \caption{Number of examples per data augmentation step}
    \label{examples_per_data_augmentation_type}
\end{figure}

After the augmentation process, the training data contains 45,514 examples (23,082 positive examples and 22,432 negative ones). Figure~\ref{examples_per_data_augmentation_type} shows the growth of the training dataset after each data augmentation step.

\subsection{Model Structure}
\label{sec:model_structure}

\begin{figure*}
    \centering
    \includegraphics[width=0.65\textwidth]{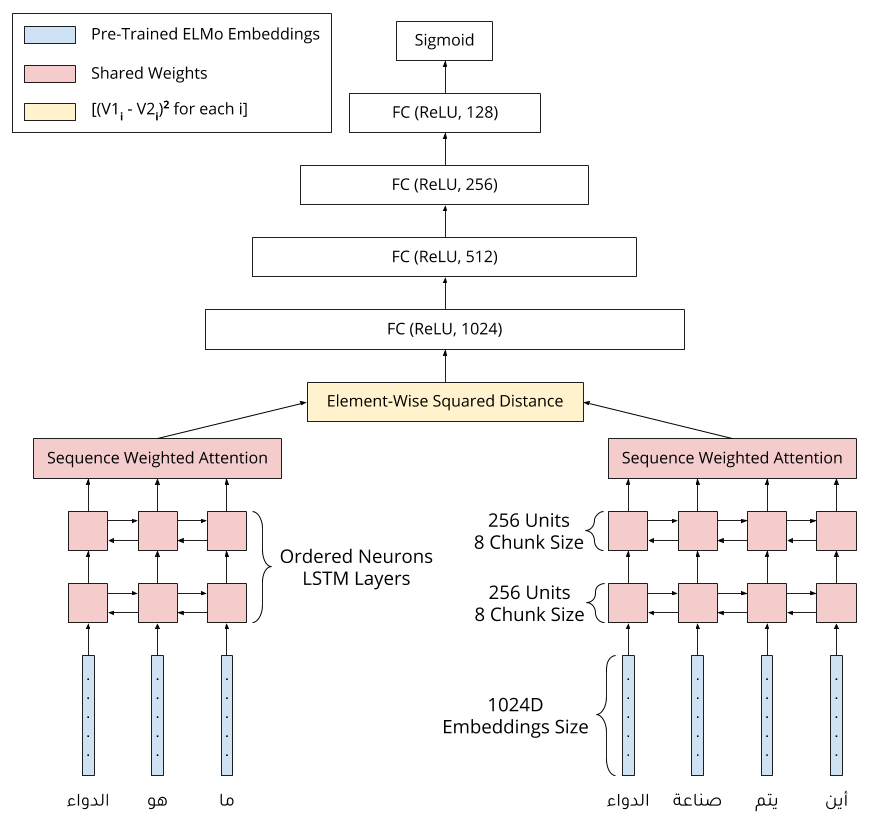}
    \caption{Model Structure}
    \label{model_representation}
\end{figure*}

We now discuss our model structure, which is shown in Figure~\ref{model_representation}. As the figure shows, the model structure can be divided into the following components/layers: input layer, sequence representation extraction layer, merging layer and decision layer. The following subsections explain each layer/component in details.

\subsubsection{Input}

To build meaningful representations for the input sequences, we use the Arabic ELMo pre-trained model\footnote{\url{https://github.com/HIT-SCIR/ELMoForManyLangs}}
to extract contextual words embeddings with size 1024 and feed them as input to our model. The representations extracted from the ELMo model are the averaged sum of word encoder and both first and second Long Short-Term Memory (LSTM) hidden layers. These representations are affected by the context in which they appear \cite{cheng2015contextual,peters2018deep,smith2019contextual}. For example, the word ``\<ذهب>'' will have different embedding vectors related to the following two sentences as they have different meanings (`gold' in the first sentence and `went' in the second one):
\begin{quote}
\centering
\<
ذهب علي كثير
> \\
Translation: Ali has a lot of gold. \\
\<
ذهب علي بعيدا
> \\
Translation: Ali went away.
\end{quote}

\subsubsection{Sequence Representation Extractor}

This component takes the ELMo embeddings related to each word in the question as an input and feeds them into two a special kind of bidirectional LSTM layers called Ordered Neurons LSTM (ON-LSTM)\footnote{\url{https://github.com/CyberZHG/keras-ordered-neurons}} introduced in \cite{shen2018ordered} with 256 hidden units, 20\% dropout rate, and 8 as the chunk size for each of them. Then, it applies sequence weighted attention\footnote{\url{https://github.com/CyberZHG/keras-self-attention}} proposed by \cite{felbo2017using} on the outputs of the second ON-LSTM layer to get the final question representation. This component uses the same weights to compute representations for each question in the pair. The details of this component are as follows \cite{shen2018ordered}.

Since NLP data are structured in a hierarchical manner, the authors of ON-LSTM \cite{shen2018ordered} proposed a new form of update and activation functions (in order to enforce a bias towards structuring a hierarchy of the data) to the standard LSTM model reported below:
\begin{align}
f_t &= \sigma_g (W_f x_t + U_f h_{t-1} + b_f)\\
i_t &= \sigma_g (W_i x_t + U_i h_{t-1} + b_i)\\
o_t &= \sigma_g (W_o x_t + U_o h_{t-1} + b_o)\\
\hat{c_t} &= \tanh (W_c x_t + U_c h_{t-1} + b_c)\\
h_t &= o_t \circ \tanh (c_t)
\end{align}

The newly proposed activation function is $cumax = cumsum(softmax(x))$, where $cumsum$ denotes the cumulative sum function. Among the desired properties of this function is to control the updates on the memory cell such that the higher ranking neurons get updated less frequently (storing long-term and global information) compared to the lower ranking neurons, which are updated more frequently (storing short-term and local information). This makes the neurons updates dependent on each other in contrast to the updates on the standard LSTM neurons.

The following equations define the new master input and forget gates and the new memory cell update function based on the new activation function:
\begin{align}
\tilde{f_t} &= cumax(W_{\tilde{f}} x_t + U_{\tilde{f}} h_{t-1} + b_{\tilde{f}})\\
\tilde{i_t} &= 1 - cumax(W_{\tilde{i}} x_t + U_{\tilde{i}} h_{t-1} + b_{\tilde{i}})\\
w_t &= \tilde{f_t} \circ \tilde{i_t}\\
\hat{f_t} &= f_t \circ w_t + (\tilde{f_t} - w_t)\\
\hat{i_t} &= i_t \circ w_t + (\tilde{i_t} - w_t)\\
c_t &= \hat{f_t} \circ c_{t-1} + \hat{i_t} \circ \hat{c_t}
\end{align}

The attention mechanism (inspired by \cite{bahdanau2014neural,yang2016hierarchical}) allows the model to learn to decide the importance of each word and build the final question representation vector based on important words only, while tuning out less important words. With a single parameter, $w_a$, the attention mechanism can be described as follows:
\begin{align}
e_t &= h_t w_a\\
a_t &= \frac{exp(e_t)}{\sum_{i=1}^{T} exp(e_i)}\\
v &= \sum_{i=1}^{T} a_i h_i
\end{align}
The weight matrix $w_a$ is the only new trainable parameter which learns the attention mechanism over the outputs of the second ON-LSTM layer. To calculate the importance scores, $a_t$, for each time step, it first multiplies each time step output, $h_t$, by the weight matrix, $w_a$, and normalizes the results using a Softmax function. Finally, the final sequence representation, $v$, is the weighted sum over all ON-LSTM outputs using the importance scores calculated earlier as weights.

\subsubsection{Merging Technique}

After extracting the representations related to each question, we merge them using pairwise squared distance function applied to the representation vectors of the two questions in each question pair. More formally, if $V1$ and $V2$ are these representation vectors, then, the merged representation vector $Vm$ can be expressed as follows:
\begin{align}
Vm = 
\begin{bmatrix}
(V1_{1} - V2_{1})^2 \\ (V1_{2} - V2_{2})^2 \\ \vdots \\ (V1_{512} - V2_{512})^2
\end{bmatrix}
\end{align}

This component allows for the Symmetric augmentation step (Section~\ref{sec:data_aug}) to enhance the results, since the $(A, B)$ examples are computationally different (in the back propagation step) from the $(B, A)$ examples.

\subsubsection{Deep Neural Network}
The final component is a deep neural network that consists of four fully-connected layers with 1024, 512, 256, and 128 units using ReLU activation function and 20\% dropout rate applied to each layer. This network takes the merged representation vector, $Vm$, as an input and predicts the label using a Sigmoid function as an output.

\section{Experiments and Results}
\label{sec:result}

In this section, we start by discussing our experimental setup. We then discuss all experiments conducted and provide detailed analysis of their results.

\subsection{Experimental Setup}

All experiments discussed in this work have been done on the Google Colab\footnote{\url{https://colab.research.google.com}} \cite{carneiro2018performance} environment using Tesla T4 GPU accelerator with the following hyperparameters:
\begin{itemize}
  \item Optimizer: Adam
  \item Learning Rate: 0.001
  \item Loss Function: Binary Cross Entropy
  \item Batch Size: 256
  \item Number of Epochs: 100
\end{itemize}

The experiments are divided into two sets. The first set aims to explore the effect of the Recurrent Neural Network (RNN) cell type, while the second set aims to explore the effect of the data augmentation techniques mentioned in Section~\ref{sec:data_aug}.

For each experiment, five models are trained and the following results are reported:
\begin{itemize}
  \item Minimum F1 score gained on the test set.
  \item Maximum F1 score gained on the test set.
  \item Average F1 score gained from the five trained models.
  \item Majority Voting F1 score gained by ensembling the five trained models.
\end{itemize}

\subsection{Effect of RNN Cell Type}
\label{sec:package_1}

\begin{table*}
\centering
\caption{Model size and training time for each RNN cell type}
\label{tab:p1_1}
\begin{tabular}{|c|c|c|}
\hline
RNN Cell               & \#Params & Training Time            \\ \hline
GRU                    & 4,363K   & 55.2s/epoch - 1.53 hours \\ \hline
LSTM                   & 5,413K   & 58.2s/epoch - 1.61 hours \\ \hline
ON-LSTM (Chunk: 4) & 5,938K   & 74.2s/epoch - 2.06 hours \\ \hline
ON-LSTM (Chunk: 8) & 5,675K   & 74.4s/epoch - 2.06 hours \\ \hline
\end{tabular}
\end{table*}

\begin{table*}
\centering
\caption{Model F1-score using different RNN cell types}
\label{tab:p1_2}
\begin{tabular}{|c|c|c|c|c|c|}
\hline
Leaderboard              & RNN Cell                                                              & Min             & Max             & Avg             & Vote            \\ \hline
\multirow{4}{*}{Public}  & GRU                                                                   & 94.075          & 94.793          & 94.613          & 95.242          \\ \cline{2-6} 
                         & LSTM                                                                  & 94.614          & 95.152          & 94.901          & 95.062          \\ \cline{2-6} 
                         & ON-LSTM (Chunk: 4)          & 94.524          & 95.601          & 95.242          & 96.140          \\ \cline{2-6} 
                         & \textbf{ON-LSTM (Chunk: 8)} & \textbf{95.601} & \textbf{95.780} & \textbf{95.691} & \textbf{96.499} \\ \hline
\multirow{4}{*}{Private} & GRU                                                                   & 93.271          & 94.194          & 93.855          & 94.579          \\ \cline{2-6} 
                         & LSTM                                                                  & 93.925          & 94.271          & 94.040          & 94.117          \\ \cline{2-6} 
                         & ON-LSTM (Chunk: 4)          & 93.810          & 94.425          & 94.224          & 94.732          \\ \cline{2-6} 
                         & \textbf{ON-LSTM (Chunk: 8)} & \textbf{94.002} & \textbf{94.463} & \textbf{94.309} & \textbf{94.848} \\ \hline
\end{tabular}
\end{table*}

In this experiments set, we use the same structure described in Section~\ref{sec:model_structure} while changing the RNN cell type only. We use all 45,514 examples from the augmented dataset in the training process.
The tested RNN cells are: Gated Recurrent Unit (GRU) \cite{cho2014learning}, LSTM \cite{hochreiter1997long} and ON-LSTM \cite{shen2018ordered}. The latter one is tested using two chunk sizes, 4 and 8, in order to explore the effect of chunk size on the training process and the size of the model.
Table~\ref{tab:p1_1} shows the model size in terms of trainable parameters and the training time for each RNN cell type, while Table~\ref{tab:p1_2} shows the F1-scores of the model using different RNN cells. Best results are shown in bold. The tables show that while GRU cells are the most efficient, the ON-LSTM cells (with chunk size 8) are the most effective (in terms of all considered measures).

\subsection{Effect of Data Augmentation}

In this experiments set, we use the RNN cell type that gives the best results in Section~\ref{sec:package_1} (ON-LSTM with chunk size 8) and the same model structure described in Section~\ref{sec:model_structure} to explore the effect of data augmentation steps mentioned in Section~\ref{sec:data_aug}.

The data augmentation steps have an effect on two factors, the training time and the accuracy measurement (F1-score). Table~\ref{tab:p2_1} shows the average training time over five runs for each data augmentation step.
Moreover, Table~\ref{tab:p2_2} shows the F1-scores of the trained model using different data augmentation types, best results shown in bold.

\begin{table*}
\centering
\caption{Model training time for each data augmentation step: O, T, S, and R, which stand for Original, Transitive, Symmetric, and Reflexive, respectively}
\label{tab:p2_1}
\begin{tabular}{|c|c|c|}
\hline
Data Augmentation & Examples Number & Training Time            \\ \hline
O                 & 11,997          & 20.0s/epoch - 0.55 hours \\ \hline
O+T               & 17,487          & 29.4s/epoch - 0.81 hours \\ \hline
O+T+S             & 34,974          & 57.0s/epoch - 1.58 hours \\ \hline
O+T+S+R           & 45,514          & 74.4s/epoch - 2.06 hours \\ \hline
\end{tabular}
\end{table*}

\begin{table*}
\centering
\caption{Model F1-score using different data augmentation types: O, T, S, and R, which stand for Original, Transitive, Symmetric, and Reflexive respectively}
\label{tab:p2_2}
\begin{tabular}{|c|c|c|c|c|c|}
\hline
Leaderboard              & Data Aug. & Min             & Max             & Avg             & Vote            \\ \hline
\multirow{4}{*}{Public}  & O                 & 93.626          & 94.703          & 94.200          & 94.973          \\ \cline{2-6} 
                         & O+T               & 93.177          & 94.434          & 93.877          & 94.793          \\ \cline{2-6} 
                         & O+T+S             & 94.344          & 94.793          & 94.631          & 95.421          \\ \cline{2-6} 
                         & \textbf{O+T+S+R}  & \textbf{95.601} & \textbf{95.780} & \textbf{95.691} & \textbf{96.499} \\ \hline
\multirow{4}{*}{Private} & O                 & 93.425          & 93.810          & 93.632          & 94.655          \\ \cline{2-6} 
                         & O+T               & 92.464          & 93.771          & 93.232          & 94.156          \\ \cline{2-6} 
                         & O+T+S             & 93.579          & 94.002          & 93.763          & 94.655          \\ \cline{2-6} 
                         & \textbf{O+T+S+R}  & \textbf{94.002} & \textbf{94.463} & \textbf{94.309} & \textbf{94.848} \\ \hline
\end{tabular}
\end{table*}

The tables show that each augmentation step affects the model's efficiency negatively. This is expected since each step incrementally increases the size of the dataset. On the other hand, not each increment step has a positive effect on the model's effectiveness. Such trends are worth exploring in a more exhaustive study. Finally, it is worth mentioning that the last experiments in both experiment sets are the same. So, they both have the same results.

\subsection{Other Attempts}

We test several other techniques to explore how they might affect our model. For example, using pre-trained FastText \cite{bojanowski2017enriching} embeddings as an input to our model yields worse F1-score on both public and private leaderboards with 94.254 and 93.118, respectively, compared with the ELMo contextual embeddings model. In another experiment, we use the thought vector outputted from the second ON-LSTM layer as input for the decision component. However, the sequence weighted attention gives better results by about 1 point of the F1-score. Moreover, an attempt to overcome the weakness of the Arabic ELMo model is done by translating the data to English using Google Translate\footnote{\url{https://translate.google.com}} and treating the problem as an English SQS problem instead, but the results are much worse with 88.868 and 87.504 F1-scores on public and private leaderboards, respectively. This is probably because a lot of information is lost during the translation process.

\subsection{Discussion}

\begin{figure}
    \centering
    \includegraphics[width=0.48\textwidth]{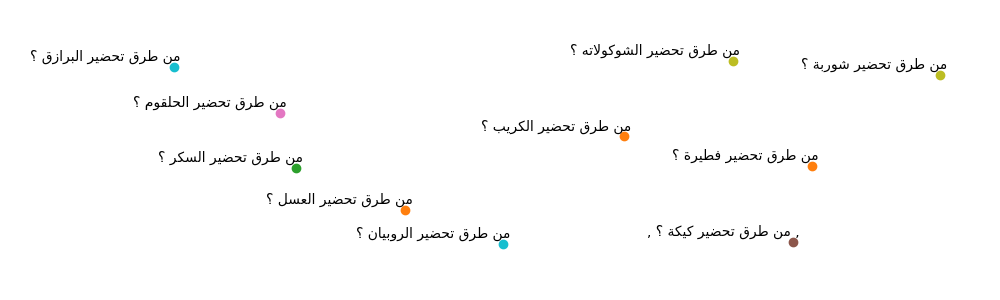}
    \caption{Representations extracted from sequence weighted attention layer for questions of the form: How to prepare `something'?}
    \label{embeddings_1}
\end{figure}

\begin{figure}
    \centering
    \includegraphics[width=0.48\textwidth]{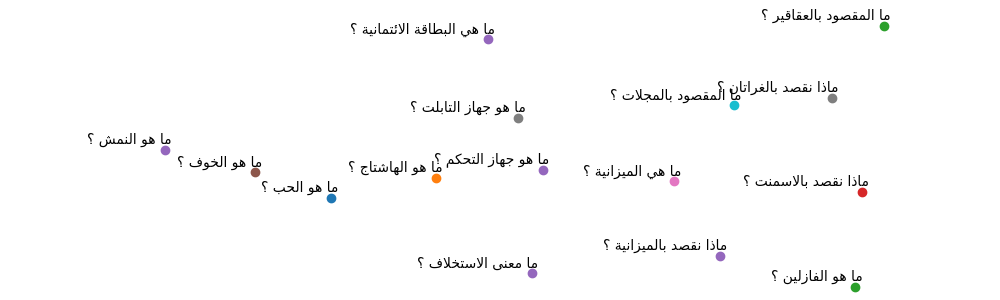}
    \caption{Representations extracted from sequence weighted attention layer for questions of the form: What is the definition of `something'?}
    \label{embeddings_2}
\end{figure}

\begin{figure}
    \centering
    \includegraphics[width=0.48\textwidth]{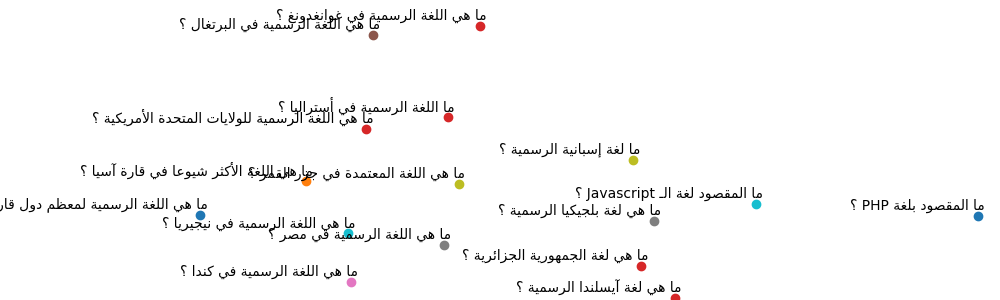}
    \caption{Representations extracted from sequence weighted attention layer for questions that ask about different language types}
    \label{embeddings_3}
\end{figure}

\begin{figure}
    \centering
    \includegraphics[width=0.48\textwidth]{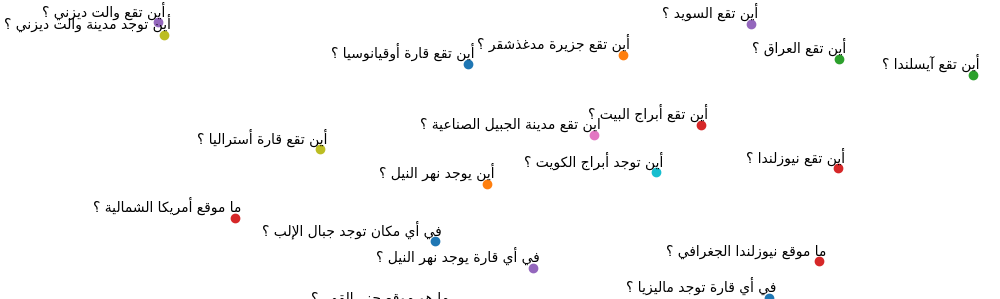}
    \caption{Representations extracted from sequence weighted attention layer for questions that ask about different places}
    \label{embeddings_4}
\end{figure}

This section briefly analyzes the questions representations learnt by our model. With the sequence weighted attention layer, the model reduces all the information about the sequence extracted using the ON-LSTMs down to a 512 fixed-size vector. By extracting these vectors from our best model and plotting them on a 2D plane using t-SNE \cite{maaten2008visualizing} dimensionality reduction algorithm, we notice some very useful observations. For example, the model learns to map questions that ask about the same thing to have nearby representations in the vector space such as the questions in Figure~\ref{embeddings_1} with the form: ``How to prepare `something'?''. The same thing goes for the questions in Figure~\ref{embeddings_2} with the form: ``What is the definition of `something'?''. In a similar manner, in Figure~\ref{embeddings_3}, the questions ask about different types of languages like ``What is the formal language in Portugal?'' and ``What is PHP language?'' are close, as well as, the questions in Figure~\ref{embeddings_4} that ask about places like ``Where is Sweden?'', ``Where is the Karak area in Jordan?'', and ``Where is the Kremlin Castle?''.

\begin{figure}
    \centering
    \includegraphics[width=0.48\textwidth]{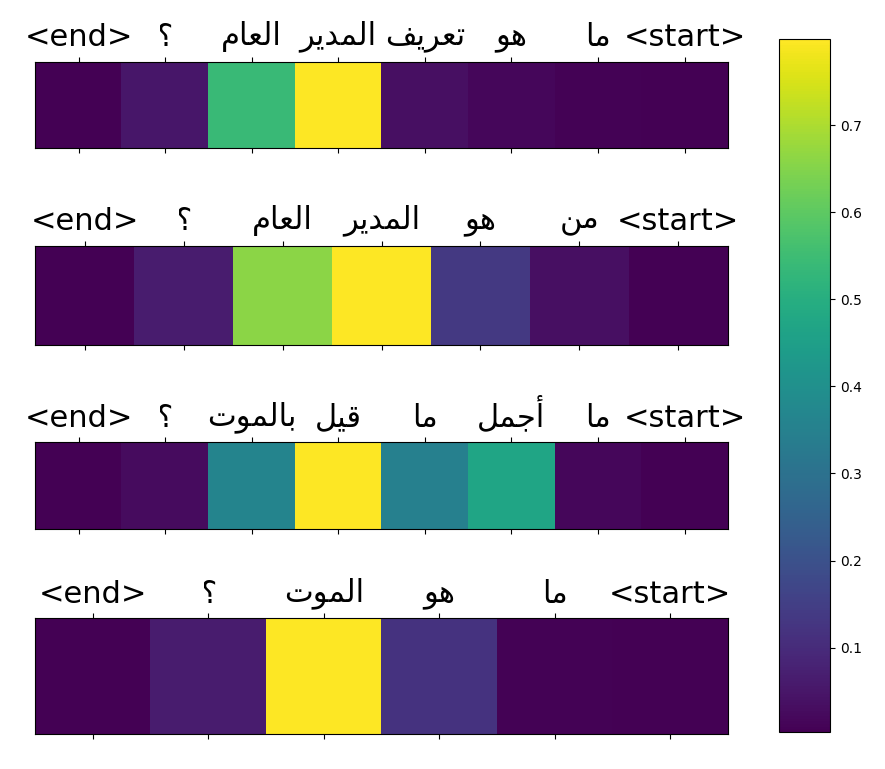}
    \caption{Weights per word from sequence weighted attention layer on four different examples}
    \label{sequence_weighted_attention}
\end{figure}

To further illustrate the usefulness of the sequence weighted attention layer, Figure~\ref{sequence_weighted_attention} shows that the attention layer learns to focus more on the key words in the questions that would determine what the question is actually asking about. This allows the model to make better decisions for whether the the questions are similar or not, even if the questions have similar words but ask about different things. The first and second questions in Figure~\ref{sequence_weighted_attention} ask about ``What is the general manager?''. So, the attention layer focuses on ``the general manager'' which is ``\<المدير العام>''. However, in the third and fourth questions, one asks ``What is the most beautiful thing that is said about death?'' and the other ones asks ``What is death?'', although both questions are related to ``death'' which is ``\<الموت>'' but the attention layer distinguishes them as not similar, where in the former one, the focus is concentrated by order on the words
``\<قيل>'',
``\<أجمل>''
and
``\<بالموت>''
(``said'', ``most beautiful'' and ``death''), while the latter one focuses mostly on
``\<الموت>''
(``death'').

\section{Conclusion}
\label{sec:conc}
In this paper, we described our team's effort on the semantic text question similarity task of NSURL 2019. Our top performing system utilizes several innovative data augmentation techniques to enlarge the training data. Then, it takes ELMo pre-trained contextual embeddings as an input and builds sequence representation vectors that are used to predict the relation between the question pairs. The model was ranked in the 1st place with 96.499 F1-score (same as the second place F1-score) and the 2nd place with 94.848 F1-score (differs by 1.076 F1-score from the first place) on the public and private leaderboards, respectively.

\section*{Acknowledgments}
We gratefully acknowledge the support of the Deanship of Research at the Jordan University of Science and Technology for supporting this work via Grant \#20180193.

\bibliography{acl2019}
\bibliographystyle{acl_natbib}

\end{document}